\documentclass[sigconf,9.5pt]{acmart}

\usepackage{hyperref}       
\usepackage{url}            
\usepackage{booktabs}       
\usepackage{amsfonts}       
\usepackage{nicefrac}       
\usepackage{amsmath}
\usepackage{graphicx}
\usepackage{float}

\newcommand{\parta}{\paragraph{Zero-Knowledge Attack Evaluation.}}
\newcommand{\partb}{\paragraph{Perfect-Knowledge Attack Evaluation.}}
\newcommand{\partc}{\paragraph{Limited-Knowledge Attack Evaluation.}}
\newcommand{\numdefenses}{ten }
\newcommand{\numpapers}{seven }
\newcommand{\Numdefenses}{Ten }

\acmConference[Submission to ACM CCS 2017]{ACM Conference on Computer and Communications Security}{Due 19 May 2017}{Dallas, Texas}
\acmYear{2017}

\copyrightyear{2017} 
\acmYear{2017} 
\setcopyright{acmlicensed}
\acmConference{AISec'17}{November 3, 2017}{Dallas, TX, USA}\acmPrice{15.00}\acmDOI{10.1145/3128572.3140444}
\acmISBN{978-1-4503-5202-4/17/11}

\title{Adversarial Examples Are Not Easily Detected: \\
  Bypassing \Numdefenses Detection Methods}
\date{}
\author{Nicholas Carlini \qquad David Wagner  \\ University of California, Berkeley}

\begin{document}

\begin{abstract}
  Neural networks are known to be vulnerable to
  adversarial examples: inputs that are close
  to natural inputs but classified incorrectly.
  In order to better understand the space of adversarial examples,
  we survey \numdefenses recent proposals that
  are designed for \emph{detection} and compare their efficacy.
  We show that \emph{all} can be defeated by constructing new loss
  functions.
  We conclude that adversarial examples are significantly harder to detect
  than previously appreciated, and the properties believed to be intrinsic
  to adversarial examples are in fact not. Finally, we propose several
  simple guidelines for evaluating future proposed defenses.
\end{abstract}

\maketitle


\section{Introduction}

Recent years have seen rapid growth in the area of machine learning. Neural networks,
an idea that dates back decades, have been a driving force behind this
rapid advancement.
Their successes have been demonstrated in a wide set of domains, from classifying images
\cite{szegedy2015rethinking}, to beating humans at Go \cite{silver2016mastering},
to NLP \cite{wu2016google,petrov2016announcing},
to self driving cars \cite{bojarski2016end}.

In this paper, we study neural networks applied to image classification.
While neural networks are the most accurate machine learning approach
known to date,
they are against an adversary who attempts to fool the classifier
\cite{biggio2013evasion}.
That is, given a natural image $x$, an adversary can easily produce a visually
similar image $x'$ that has a different classification. Such
an instance $x'$ is known as an \emph{adversarial example} \cite{szegedy2013intriguing},
and they have been shown to exist
in nearly all domains that neural networks are used.

The research community has reacted to this observation in force, proposing many
defenses that attempt to classify adversarial examples correctly
\cite{gu2014towards,jin2015robust,papernot2016distillation,zheng2016improving,
  rozsa2016adversarial,huang2015learning,shaham2015understanding,bastani2016measuring}.
Unfortunately, most of these defenses are not effective at classifying adversarial
examples correctly. 

Due to this difficulty, recent work has turned to
attempting to \emph{detect} them instead.
We study \numdefenses detection
schemes proposed in \numpapers papers over the last year
\cite{hendrik2017detecting,li2016adversarial,grosse2017statistical,hendrycks2017early,
  feinman2017detecting,bhagoji2017dim,gong2017adversarial},
and compare their efficacy with the other defenses in a consistent manner.
With new attacks, we show that in every case the defense can be evaded by an
adversary who targets that specific defense. On simple datasets, the
attacks slightly increase the distortion required, but on more complex
datasets, adversarial examples remain completely indistinguishable
from the original images.

By studying these recent schemes that detect adversarial examples,
we challenge the assumption that adversarial examples have
intrinsic differences from natural images. We also use these experiments to
obtain a better understanding of the space of adversarial examples.

We evaluate these defenses under three threat models.
We first consider a
generic attacks that don't take any specific measures to fool any
particular detector.
We show six of the \numdefenses defenses are significantly less effective
than believed under this threat model.
Second, we introduce novel white-box attacks that break each defense when
tailored to the given defense; five of the defenses provide \emph{no}
increase in robustness;
three increase robustness only slightly; the final two
increase effective only on simple datasets.
Our attacks work by defining
a special attacker-loss function that
captures the requirement that the adversarial examples must fool
the defense, and optimizing for this loss function.
We discover that the specific loss function chosen is
critical to effectively defeating the defense: choosing the immediately obvious
loss function often results in the defense appearing significantly more robust
than it actually is.
Finally, we leverage the
transferability~\cite{szegedy2013intriguing} property to work even when the
adversary does not have knowledge of the defense's model parameters.


Our results further suggest that there is a need for
better ways to evaluate potential defenses.
We believe our approach would be a useful baseline: to be worth
considering, a proposed defense should follow the approach used here as a
first step towards arguing robustness.

The code to reproduce our results is available online at \\
\mbox{\url{http://nicholas.carlini.com/code/nn_breaking_detection}}.

We make the following contributions:
\begin{itemize}
\item We find that many defenses are unable to detect
  adversarial examples,
  even when the attacker is oblivious to the specific defense used.
\item We break all existing detection methods in the white-box (and
  black-box) setting
  by showing how to pick good attacker-loss functions for each defense.
\item We draw conclusions about the space of adversarial examples, and
  offer a note of caution about evaluating solely on MNIST; it appears that
  MNIST has somewhat different security properties than CIFAR.
\item We provide recommendations for evaluating defenses.
\end{itemize}

\section{Background}

The remainder of this section contains a brief survey of the field of neural networks and
adversarial machine learning. We encourage readers unfamiliar with this area to read
the following papers (in this order): \cite{szegedy2013intriguing},
\cite{goodfellow2014explaining},
\cite{papernot2016transferability}, and \cite{carlini2016towards}.

\subsection{Notation}
\label{sec:notation}
Let $F(\cdot)$ denote a neural network
used for classification. The final layer in this network is a softmax
activation, so that the output is a probability distribution where $F(x)_i$ represents
the probability that object $x$ is labeled with class $i$.

All neural networks we study are feed-forward networks consisting of multiple layers
$F^i$ taking as input the result of previous layers.
The outputs of the final layer are known as logits;
we represent them by $Z(\cdot)$. Some layers involve the non-linear
ReLU \cite{nair2010rectified} activation.
Thus the $i$th layer computes
\[ F^i(x) = \text{ReLU}(A^i \cdot F^{i-1}(x) + b^i) \]
where $A^i$ is a matrix and $b^i$ is a vector.
Let $Z(x)$ denote the output of the last layer (before the softmax),
i.e., $Z(x)=F^n(x)$.
Then the final output of the network is
\[ F(x) = \text{softmax}(Z(x)). \]
When we write $C(x)$ we mean the classification of $F(\cdot)$ on $x$:
\[ C(x) = \text{arg max}_i(F(x)_i). \]

Along with the neural network, we are given a set of training instances with
their corresponding labels $(x,y) \in \mathcal{X}$.

\subsection{Adversarial Examples}

The security of machine learning is a well studied field: early
work considered this problem mostly on linear classifiers
\cite{dalvi2004adversarial,lowd2005adversarial};
later work more generally examined the security of machine learning
\cite{barreno2006can,barreno2010security} to both evasion and poising attacks.

More recently, Biggio \emph{et al.}
and Szegedy \emph{et al.} \cite{biggio2013evasion,szegedy2013intriguing}
demonstrated test-time evasion attacks on neural networks. They were able to
produce visually similar images that had different labels assigned by
the classifier.

We begin by defining
an input to the classifier $F(\cdot)$ \emph{natural} if it is an instance
that was benignly created (e.g., all instances in the training set
and testing set are natural instances).
Then, given a network $F(\cdot)$ and a natural input $x$ so that $C(x) = l$ we say
that $x'$ is an (untargeted) \emph{adversarial example} if $x'$ is close to
$x$ and
$C(x') \ne l$. A more restrictive case is where the adversary picks a target
$t \ne l$ and seeks to find $x'$ close to $x$ such that $C(x')=t$; in this case we
call $x'$ a \emph{targeted} adversarial example. 
We focus on targeted adversarial examples
exclusively in this paper.
When we say a neural network is \emph{robust} we mean
that it is difficult to find adversarial examples on it.

To define closeness, most attacks use an $L_p$ distance, defined as
$\| d \|_p = (\sum_{i=0}^n |v_i|^p)^{1 \over p}$.
Common choices of $p$ include: $L_0$, a measure of the number of pixels changed
\cite{papernot2016limitations};
$L_2$, the standard Euclidean norm \cite{szegedy2013intriguing,carlini2016towards,moosavi2016deepfool};
or $L_\infty$, a measure of the
maximum absolute change to any pixel \cite{goodfellow2014explaining}.
If the total distortion under any of these three distance metrics is small,
the images will likely appear visually similar.
We quantitatively measure
the robustness of a defense in this paper by measuring the distance to the nearest
adversarial example under the $L_2$ metric.

One further property of adversarial examples we will make use of is
the transferability property \cite{szegedy2013intriguing,goodfellow2014explaining}.
It is often the case that, when given two models $F(\cdot)$ and $G(\cdot)$, an
adversarial example on $F$ will also be an adversarial example on $G$, even if they
are trained in completely different manners, on completely different training sets.

There has been a significant amount of work studying methods to construct
adversarial examples \cite{szegedy2013intriguing,biggio2013evasion,
  goodfellow2014explaining,papernot2016limitations,moosavi2016deepfool,
  carlini2016towards}
and to make networks robust against adversarial examples
\cite{gu2014towards,jin2015robust,papernot2016distillation,zheng2016improving,
  rozsa2016adversarial,huang2015learning,shaham2015understanding,bastani2016measuring}.
To date, no defenses has been able to classify adversarial examples correctly.

Given this difficulty in correctly classifying adversarial examples, recent defenses have
instead turned to detecting adversarial examples and reject them. We study
these defenses in this paper \cite{hendrik2017detecting, gong2017adversarial, grosse2017statistical, feinman2017detecting, li2016adversarial, bhagoji2017dim, hendrycks2017early}.


\subsection{Threat Model}

As done in Biggio \emph{et al.} \cite{biggio2013evasion}, we consider three different threat models in this paper:
\begin{enumerate}
\item An \emph{Zero-Knowledge Adversary} generates adversarial examples on the
  unsecured model $F$ and is not aware that the detector $D$ is in place. The
  detector is successful if it can detect these adversarial examples.
\item A \emph{Perfect-Knowledge Adversary} is aware the neural network is
  being secured with a given detection scheme $D$, knows the model parameters
  used by $D$, and can use these to attempt to evade both the
  original network $F$ and the detector simultaneously.
\item A \emph{Limited-Knowledge Adversary} is aware the neural network is being
  secured with a given detection scheme, knows how it was trained, but
  does not have access to the trained detector $D$ (or the exact training data).
\end{enumerate}
We evaluate each defense under these three threat models. We discuss our
evaluation technique in Section~\ref{sec:howeval}.




\subsection{Datasets}

In this paper we consider two datasets used throughout the existing work in this field.

The \emph{MNIST} dataset \cite{lecun1998mnist} consists of $70,000$
  $28\times28$ greyscale images of handwritten digits from 0 to 9.
  Our standard convolutional network achieves $99.4\%$ accuracy on this dataset.

The \emph{CIFAR-10} dataset \cite{krizhevsky2009learning} consists of $60,000$
  $32\times32$ color images of ten different objects (e.g., truck, airplane, etc).
This dataset is substantially more difficult: the
state of the art approaches achieve $95\%$ accuracy \cite{springenberg2014striving}.
For comparison with prior work, we use the ResNet \cite{he2016deep} architecture
from Metzen \emph{et al.} \cite{hendrik2017detecting} trained in the same manner.
This model achieves a $91.5\%$ accuracy.

The first row of Figure~\ref{fig:show_mnistcifar} shows natural examples drawn
from the test set of these datasets.


\subsection{Defenses}

In order to better understand what properties are intrinsic of adversarial
examples and what properties are only artificially true because of existing
attack techniques, we choose the first seven papers released that
construct defenses to detect adversarial examples.

Three of the defenses \cite{grosse2017statistical,
  gong2017adversarial,hendrik2017detecting} use a second neural network to
classify images as natural or adversarial. Three use PCA to detect
statistical properties of the images or network paramaters \cite{li2016adversarial,
  hendrycks2017early,bhagoji2017dim}. Two perform other statistical tests
\cite{grosse2017statistical,feinman2017detecting}, and the final two
perform input-normalization with randomization and blurring \cite{feinman2017detecting,
  li2016adversarial}.

\begin{figure*}
\small
  \begin{minipage}{0.07\textwidth}\raggedright
    Reference\\
    \vspace{.45cm}
    Unsecured\\
    \vspace{.45cm}
    Grosse \\ \S\ref{sec:retrain}\\
    \vspace{.08cm}
    Gong \\ \S\ref{sec:retrain}\\
    \vspace{.08cm}
    Metzen \\ \S\ref{sec:secondary}\\
    \vspace{.08cm}
    Hendrycks \\ \S\ref{sec:inputpca}\\
    \vspace{.08cm}
    Bhagoji \\ \S\ref{sec:dimreduction}\\
    \vspace{.08cm}
    Li \\ \S\ref{sec:pcacnn}\\
    \vspace{.08cm}
    Grosse \\ \S\ref{sec:mmd}\\
    \vspace{.08cm}
    Feinman \\ \S\ref{sec:de}\\
    \vspace{.08cm}
    Feinman \\ \S\ref{sec:dropout}\\
    \vspace{.08cm}
    Li \\ \S\ref{sec:blur}\\
    \vspace{.6cm}
  \end{minipage}  
  \begin{minipage}{0.89\textwidth}
  \includegraphics[scale=0.07]{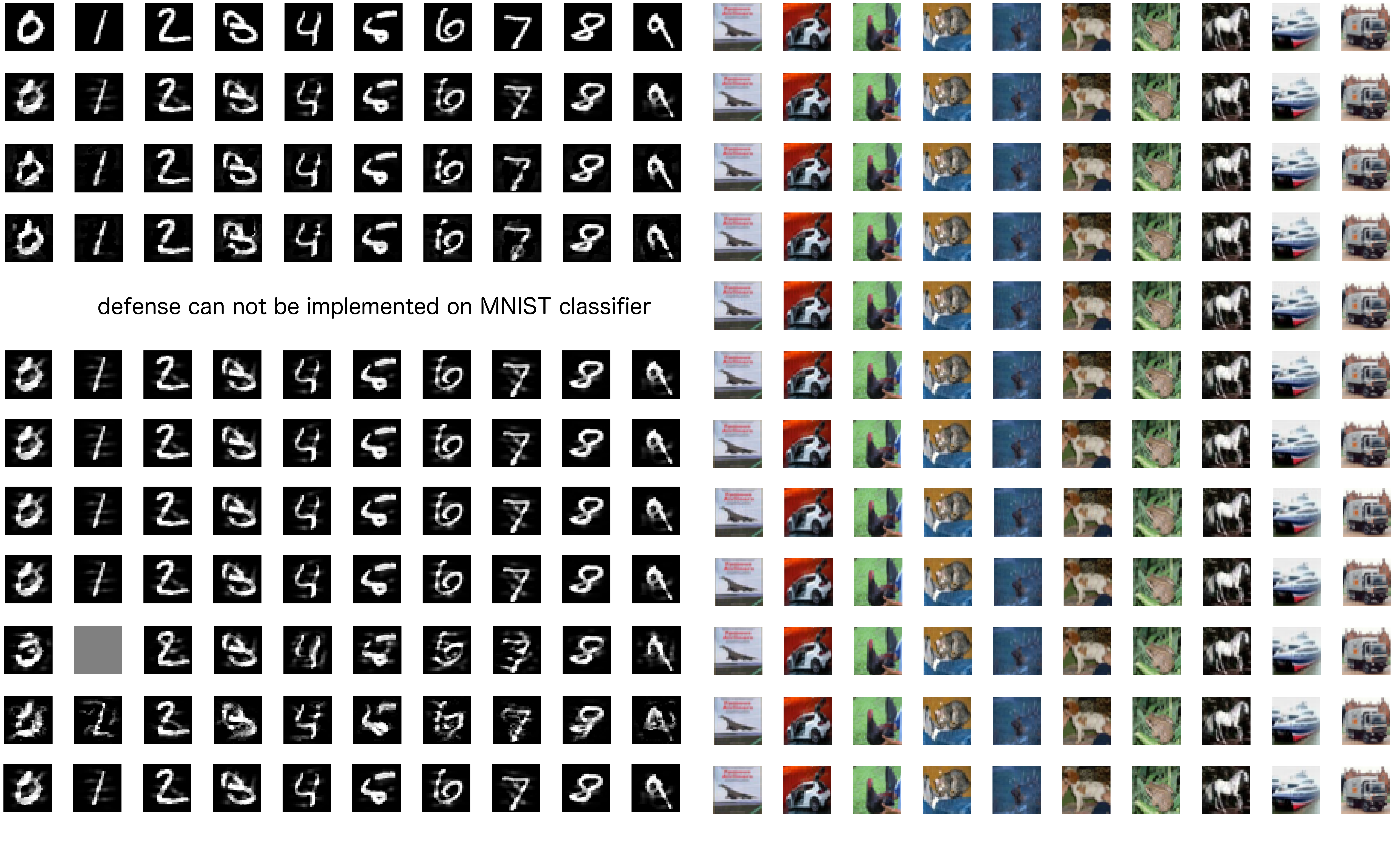}
  \end{minipage}
  \caption{\textbf{Summary of Results}: adversarial examples on the MNIST and CIFAR datasets for each defense
    we study. The first row corresponds to the original images.}
  \label{fig:show_mnistcifar}
\end{figure*}

We summarize our results in Figure~\ref{fig:show_mnistcifar}. Some defenses can slightly
increase distortion required for MNIST digits. However, no defense makes CIFAR
adversarial examples visually distinguishable from the original image. We generate
adversarial examples as described below.

\subsection{Generating Adversarial Examples}

We use the $L_2$ attack algorithm of Carlini and Wagner \cite{carlini2016towards} to
generate targeted adversarial examples,
as it is superior to other published attacks.
At a high level it is an iterative attack as done in the initial work on
constructing adversarial examples \cite{biggio2013evasion,szegedy2015rethinking}.
Given a neural network $F$ with logits
$Z$, the attack uses gradient descent to solve
\[ \text{minimize } \; \|x' -x\|_2^2 + c \cdot \ell(x') \]
where the loss function $\ell$ is defined as
\[ \ell(x') = \max(\max \{ Z(x')_i : i \ne t\} - Z(x')_t, -\kappa). \]

We now give some intuition behind this loss function. The difference
$\max \{ Z(x')_i : i \ne t\} - Z(x')_t$ is used to compare the target class
$t$ with the next-most-likely class. However, this is minimized when the
target class is significantly more likely than the second most likely class,
which is not a property we want. This is fixed by taking the maximum of this
quantity with $-\kappa$, which controls the confidence of the adversarial
examples. When $\kappa=0$, the adversarial examples are called \emph{low-confidence
  adversarial examples} and are only just classified as the target class.
As $\kappa$ increases, the model classifies the adversarial example as
increasingly more likely, we call these \emph{high-confidence adversarial
  examples}.

The constant $c$ is chosen via binary search. If $c$ is too small, the
distance function dominates and the optimal solution will not have a
different label. If $c$ is too large, the objective term dominates and
the adversarial example will not be nearby.

Of critical importance is that the loss function operates over the logits $Z$,
and not the probabilities $F$. As described in \cite{carlini2016towards},
the optimal choice of the constant $c \sim {1 \over |\nabla \ell|}$; therefore,
if $F$ were used instead of $Z$, no ``good'' constant $c$ would exist since
$f$ varies by several orders of magnitude (and $Z$ usually only by one).
When constructing attacks in later sections, we often choose new loss
functions $\ell$ that also do not vary in their magnitude.

Aside from C\&W's attack,
the \emph{Fast Gradient Sign} attack \cite{goodfellow2014explaining} and
\emph{JSMA} \cite{papernot2016limitations} are two attacks used by some
defenses for evaluation. These attacks are weaker than C\&W's attack and we
do not use them for evaluation \cite{carlini2016towards}.

\subsection{Attack Approach}
\label{sec:howeval}
In order to evaluate the robustness of each of the above defenses, we take
three approaches to target each of the three threat models introduced earlier.

\paragraph{Evaluate with a strong attack (Zero-Knowledge):}
In this step we generate adversarial examples with C\&W's attack
and check whether the defense can detect this strong attack.
This evaluation approach has the weakest threat model (the attacker is
not even aware the defense is in place), so any defense should trivially be able to
detect this attack.
Failing this test implies that the second two tests will also fail.

\paragraph{Perform an adaptive, white-box attack (Perfect-Knowledge):}
The most powerful threat model, we assume here the adversary has access to the
detector and can mount an adaptive attack. To perform this attack, we construct a
new loss function, and generate
adversarial examples that both fool the classifier and
also evade the detector.

The most difficult step in this attack is to construct a loss
function that can be used to generate adversarial examples. In some cases, such
a loss function might not be readily available.
In other cases, one may exist, but it may not be
well-suited to performing gradient descent over. It is of critical importance
to choose a good loss function, and we describe how to construct such a loss
function for each attack.

\paragraph{Construct a black-box attack (Limited-Knowledge):}
This attack is 
the most difficult for the adversary. We assume the
adversary knows what type of defense is in place but does not know the detector's
paramaters.
This evaluation is only interesting if (a) the zero-knowledge attack failed to generate
adversarial examples, and (b) the perfect-knowledge attack succeeded. If the strong
attack alone succeeded, when the adversary was not aware of the defense,
they could mount the same attack in this black-box case. Conversely, if the white-box
attack failed, then a black-box attack will also fail (since the threat model
is strictly harder).

In order to mount this attack, we rely on the transferability property:
the attacker trains a substitute model in the same way as the original
model, but on a separate training set (of similar size, and quality).
The attacker can access substitute model's parameters, and performs a white-box attack
on the substitute model. Finally,
we evaluate whether these adversarial examples transfer
to the original model.

When the classifier and detector are separate models, we assume the adversary has
access to the classifier but not the detector (we are analyzing the increase in
security by using the detector).

If the detector and classifier are not separable (i.e., the classifier is trained
to also act as a detector), then to perform a fair evaluation,
we compare
the adversarial examples generated with black-box access to the (unsecured) classifier
to adversarial examples generated with only black-box access to both the classifier and
detector.


\section{Secondary Classification Based Detection}
We now turn to evaluating the ten defenses.
The first category of detection schemes we study build a second classifier
which attempts to detect adversarial examples.
Three of the approaches take this direction.

For the remainder of this subsection, define $F(\cdot)$ to be the classification
network and $D(\cdot)$ to be the detection network. $F(\cdot)$ is defined as
in Section~\ref{sec:notation} outputting a probability distribution over the 10 classes, and
$D : \mathbb{R}^{w \cdot h \cdot c} \to (-\infty,\infty)$ represent the logits of the
likelihood the instance is adversarial. That is, $\text{sigmoid}(D(x)) : \mathbb{R}^{w \cdot h \cdot c} \to [0,1]$ represents the probability the instance is adversarial.

\subsection{Adversarial Retraining}
\label{sec:retrain}

Grosse \emph{et al.} \cite{grosse2017statistical} propose a
variant on adversarial re-training.
Instead of attempting to classify the adversarial examples correctly (by
adding adversarial examples to the training set, with their correct labels),
they introduce a new $N+1$st class --- solely for adversarial examples --- and
train the network to detect adversarial examples.
Specifically, they propose the following procedure:
\begin{enumerate}
\item Train a model $F_{base}$ on the training data $\mathcal{X}_0 = \mathcal{X}$.
\item Generate adversarial examples on model $F_\text{base}$ for each $(x_i,y_i) \in X$. Call these
  examples $x'_i$.
\item Let $\mathcal{X}_1 = \mathcal{X}_0 \cup \{(x'_i, N+1) : i \in |\mathcal{X}|\}$ where $N+1$ is the new
  label for adversarial examples.
\item Train a model $F_\text{secured}$ on the training data $\mathcal{X}_1$.
\end{enumerate}

Gong \emph{et al.} \cite{gong2017adversarial}
construct a very similar defense technique. Instead of re-training the model
$F_\text{secured}$ completely, they construct a binary classifier $D$ that simply
learns to partitions the instances $x$ from $x'$ by training on the training set
\[ \mathcal{X}_1 = \{(x_i,1)  : i \in |\mathcal{X}|\} \cup \{(x'_i, 0) : i \in |\mathcal{X}|\}. \]


We re-implement these two defenses and find that adversarial retraining is able to
detect adversarial examples when generated with the fast gradient sign and JSMA attacks
with near-$100\%$ accuracy.

\parta
We train these two schemes on the entire MNIST training set,
using C\&W's attack to generate adversarial examples.
In this way we construct a model $F_\text{secured}$ and $D$.

We then construct adversarial examples for $F_\text{base}$
from each image in the test set using C\&W's attack.
Both approaches
detect these previously unseen test adversarial examples. Grosse \emph{et al.}
detects $98.5\%$ of attacks as adversarial. Further, it classifies half of the remaining
$1.5\%$ correctly. Gong \emph{et al.} achieve $98\%$ accuracy in detecting adversarial
examples.

Investigating further, we find that even if we train on
adversarial examples generated using an \emph{untargeted} attack, both schemes
can detect \emph{targeted} adversarial examples.

Neither of these defenses claim robustness against
CIFAR, however when we perform this evaluation, we find
to achieve a $70\%$ detection rate requires a $40\%$ false positive rate. This alone
can be seen as a failure of these approaches on CIFAR.

\partb
Next, we evaluate these defenses assuming the adversary is aware of these defenses and parameters. That is, we directly attack the defended model.
Our experiments revealed that these defenses are ineffective and add
almost no increase in robustness.

For Grosse's defense, we use C\&W's attack on $F_\text{secured}$
to generate adversarial examples; it succeeds 100\% of the time.
We computed the mean $L_2$-distance from the original sample to the
adversarial example.
Adversarial examples against $F_\text{base}$ are at average $L_2$ distance
of $2.05$ from the original sample;
adversarial examples against $F_\text{secured}$ have an average
distance of $2.26$.
Thus the defense has not reduced the success rate at generating adversarial
examples, and has only increased the mean distortion by 10\%.

Gong's defense does not fare any better.
To help construct adversarial examples that will simultaneously fool
$F_\text{base}$ and $D$, we define a new function $G(\cdot)$ that represents
the combination of the classifier (with logits $Z_F(\cdot)$)
and detector (with logits $Z_D(\cdot)$).
In particular, we define
\[ G(x)_i = \begin{cases}
  Z_F(x)_i & \text{if } i \le N \\
  (Z_D(x)+1) \cdot \max\limits_j Z_F(x)_j & \text{if } i = N+1
\end{cases} \]
Effectively, $G$ acts as a classifier on $N+1$ classes.
It is constructed to have two useful properties:
if $Z_D(x) > 0$ (i.e., if the detector classifies $x$ as malicious)
then we will have
\[\text{arg max}_i(G(x)_i)=N+1\]
(where $N$ is the new adversarial class), and if
$Z_D(x) < 0$ (i.e., if the detector classifies $x$ as natural)
then we will have
\[ \text{arg max}_i(G(x)_i) = \text{arg max}_i(Z_F(x)_i).\]

\paragraph{Why did we choose this particular function $G(\cdot)$?}
Recall from earlier that when using a
gradient-descent based attack algorithm, there is a constant $c$
in the loss function that trades
off the relative importance of reducing the distance from the natural
instance vs. classifying the input as the target class.
If the value of the
loss function changes by several orders of magnitude, no value of $c$ will be
optimal and adversarial examples will be very hard to find. Therefore,
we construct $G$ so that it is mostly linear.

Now we treat this function $G$ as if it were a neural network, and feed it
directly into C\&W's attack algorithm instead of the function $Z$. It is easy to
see that if an instance $x'$ is classified by $G$ as label $l \le N$ then both
$C(x')=l$ and the detector classifies $x'$ as natural (non-adversarial).
This approach successfully generates adversarial examples against Gong's
defense with 100\% success.
The mean $L_2$ distortion of these adversarial examples
is $2.31$, an improvement of about $10\%$ over the unprotected scheme.

We therefore conclude neither of these schemes are effective on MNIST.
While adversarial examples are detectable with neural networks
when not attempting to be evasive, it is possible to construct evasive
adversarial examples.

On CIFAR, the above attacks work in an identical manner and give
very similar results: the distortion is increased less than $5\%$.

\partc
For brevity, we only describe the attack on
Grosse's scheme (Gong's results are identical). We generate transferable adversarial
examples as described above.
Instead of training one model on the complete training set,
we split the training set in half.
We assume the first half is known to the attacker and the
second half is used by the defender to train their network.
Thus, we construct two models $R_1$ and $R_2$ each trained
on half of the training data.
We treat $R_2$ as the defender's model (the target model for which we want to
produce adversarial examples) and give the attacker white-box access to all
parameters of model $R_1$.
Then we apply the C\&W attack to each instance in the test set
to generate adversarial examples for $R_1$,
and we test whether they fool $R_2$.

As a baseline, we applied this procedure to the unsecured model.
The attack succeeded $98\%$ of the time, with a mean $L_2$ distortion
of $5.1$.
Then, we generated two secured models $R_1$ and $R_2$ implementing
Grosse's defense, and applied the same attack.
We observed a $98\%$ success rate, with a mean $L_2$ distortion of
$5.3$, which is only $4\%$ higher than the baseline.
Therefore, we conclude that Grosse's defense is ineffective and can
be attacked even by an attacker who does not the exact model parameters
of the secured model.

\subsection{Examining Convolutional Layers}
\label{sec:secondary}
           
In contrast to the prior approach, which attempts to detect adversarial
examples based on the contents of the image itself,
Metzen \emph{et al.} \cite{hendrik2017detecting} detect adversarial examples by looking at the inner
convolutional layers of the network.
They augment the classification neural network with a detection neural
network that takes its input from various intermediate layers of the
classification network.
This detection network is trained identically to Gong's defense above.
We refer interested readers to the
original paper for complete details on the detector setup  \cite{hendrik2017detecting}.
This defense only argues robustness against CIFAR, since it looks at the
inner layers of a ResNet (which are not usually used for MNIST).

We then train the detector by attaching it
to the output of the first residual block as done in \cite{hendrik2017detecting}.
We
confirmed that their detector is able to detect adversarial examples
with $99\%$ probability on simple attacks (fast gradient sign or JSMA).

\parta 
We found training the detector to be very difficult and highly sensitive;
we experimented with different optimizers, hyper-parameters,
and initializers to reduce the effect of this.

For evaluation, we generate adversarial examples for each instance
in the test set, using the C\&W attack.
The best detector we were able to train correctly obtains an $81\%$ true positive
rate at $28\%$ false positive rate. This is in stark contrast to the $99\%$
success at detecting simpler attacks, which highlights that
weak attacks give a biased view of defense robustness.

Similar to the prior defense (that also had a high false positive rate),
this defense is unable
to perform well even when the adversary is not attempting to evade it.



\partb
Our white-box attack completely defeats Metzen's defense:
it is able to produce adversarial examples that simultaneously are
mis-classified by the original network and evade the detector.
We generate adversarial examples using C\&W's attack applied to
the same function $G(\cdot)$ defined in Section~\ref{sec:retrain}.
The mean distance to
adversarial examples increases from $0.169$ $L_2$ distortion on the unsecured
model to $0.227$ on the secured scheme, an improvement of
$34\%$. However, in absolute terms, the adversarial examples generated are
still indistinguishable from the original inputs.

\partc
To investigate if this defense is robust to attacks in a black-box setting,
we perform a standard transferability test as done above. We split the training data in half,
and train two detector models, one on each half of the training data. Then, we
attack the second detector given only white-box access to the first detector.

On MNIST,
we found that even low-confidence adversarial examples transfer $84\%$ of the
time between the two detectors when the classifier network is known by the adversary.
By using high-confidence adversarial examples,
the attack success rate can be increased to $98\%$ at
the cost of increasing the mean distortion by a further $28\%$, which is
small enough that adversarial examples remain indistinguishable from
the original images.

\section{Principal Component Analysis Detection}

Principal Component Analysis (PCA) transforms a set of points in a $n$-dimensional
space to a new set of points in a $k$-dimensional space ($k \le n$) through a
linear transformation. We assume the reader is familiar with PCA for the
remainder of this section.

\subsection{Input Image PCA}
\label{sec:inputpca}

Hendrycks \& Gimpel \cite{hendrycks2017early} use PCA to detect natural images from adversarial
examples, finding that adversarial examples place a higher weight on the larger
principal components than natural images (and lower weight on the earlier principal
components).

\begin{figure}
  \centering
  \includegraphics[scale=.85]{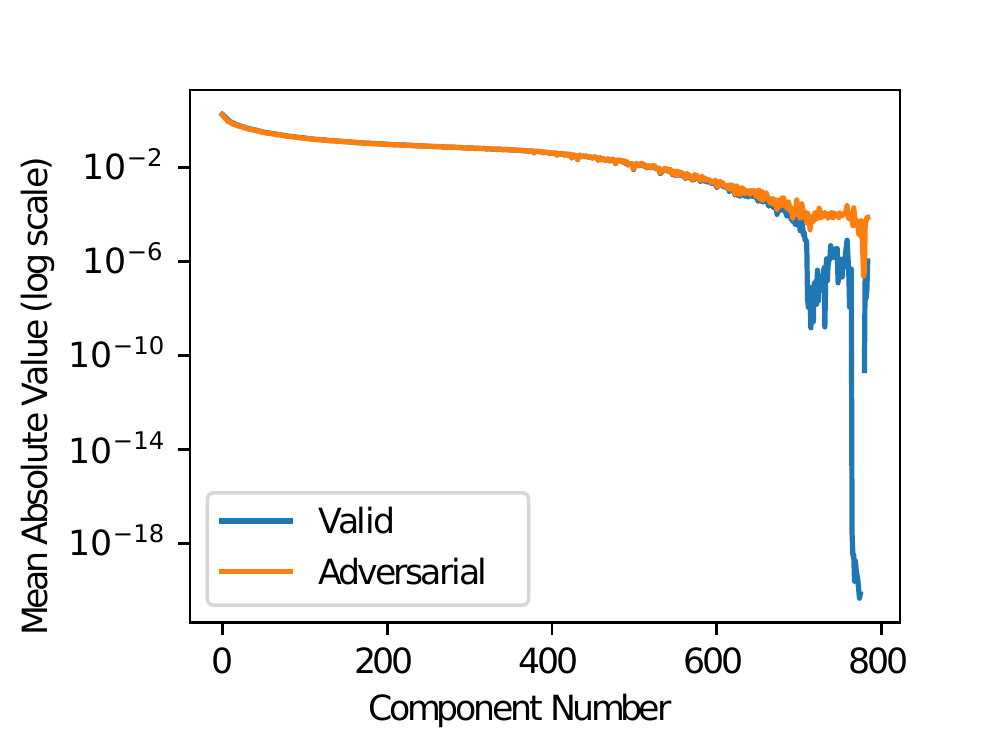}
  \caption{PCA on the MNIST dataset reveals a difference between natural images
    and adversarial images, however this is caused by an artifact of MNIST:
    border pixels on natural images are often 0 but slightly-positive on adversarial examples.}
  \label{fig:pcamnist}
\end{figure}

\parta
We first reproduce their results by running PCA on MNIST.
To see if adversarial examples really do use larger principal components more
often, we compute how much each component is used.
Let $X_1,\dots,X_n$ be the training set instances.
We define the score $S(j)$ of the $j$th PCA component as
\[S(j) = {1 \over N}\sum_{i=1}^N |PCA(X_i)_j|. \]
We train a classification network on the training set and
compute the component scores $S(1),\dots,S(784)$.
Then, for each image in the test set, we find the nearest adversarial example with
C\&W's attack and we compute the component scores on these adversarial
examples.
The results are plotted in Figure~\ref{fig:pcamnist}.

Our results agree with Hendrycks \emph{et. al} \cite{hendrycks2017early}:
there is no difference on the first principal components, but there is a
substantial difference between natural and adversarial instances on the later
components.
On the MNIST data set, their defense does detect zero-knowledge attacks,
if the attacker does not attempt to defeat the defense.

\paragraph{Looking Deeper.}
At first glance, this might lead us to believe
that PCA is a powerful and effective method for detecting adversarial examples.
However,
whenever there are large abnormalities in the data, one must be careful to understand
their cause.

In this case, the reason for the difference is that there are pixels
on the MNIST dataset that are almost always set to 0.
Since the MNIST dataset
is constructed by taking 24x24 images and centering them (by center-of-mass)  on a 28x28 grid,
the majority of the pixels on the boundary of natural images are zero.
Because these border pixels are essentially always zero for natural
instances, the last principal components are heavily concentrated on
these border pixels.
This explains why the last 74 principal components ($9.4\%$ of the components)
contribute less than $10^{-30}$ of the variance on the training set.

In short, the detected difference between the natural and adversarial examples
is because the border pixels are nearly always zero for natural MNIST instances,
whereas typical adversarial examples have non-zero values on the border.
While adversarial examples are different from natural images on
MNIST in this way, this
is not an intrinsic property of adversarial examples; it is instead due to an
artifact of the MNIST dataset.
When we perform the above evaluation on CIFAR, there is no detectable difference
between adversarial examples and natural data.
As a result, the Hendrycks defense is not effective for CIFAR --- it is
specific to MNIST.
Also, this deeper understanding of why the defense works on MNIST
suggests that adaptive attacks might be able to avoid detection
by simply leaving those pixels unchanged.

\partb
We found that the Hendrycks defense can be broken by a white-box attacker
with knowledge of the defense.
Details are deferred to Section~\ref{sec:dimreduction}, where we break
a strictly stronger defense.
In particular, we found in our experiments that we can generate adversarial
examples that are restricted to change only the first $k$ principal components
(i.e., leave all later components unchanged), and these adversarial
examples that are not detected by the Hendrycks defense.

\subsection{Dimensionality Reduction}
\label{sec:dimreduction}

Bhagoji \emph{et al.} \cite{bhagoji2017dim} propose a defense based on
dimensionality reduction:
instead of training a classifier on the original training data,
they reduce the $W \cdot H \cdot C = N$-dimensional input
(e.g., 784 for MNIST) to a much smaller $K$-dimensional input (e.g., 20) and
train a classifier on this smaller input.
The classifier uses a fully-connected neural network:
PCA loses spatial locality, so a convolutional network cannot be used (we
therefore consider only MNIST).

This defense restricts the attacker so they can only manipulate
the first $K$ components: the classifier ignores other components.
If adversarial examples rely on the last principal components (as
hypothesized),
then restricting the attack to only the first $K$ principal components should
dramatically increase the required distortion to produce an adversarial
example.
We test this prediction empirically.

We reimplement their algorithm with their same model (a fully-connected network
with two hidden layers of 100 units). We train 26 models with different
values of $K$, ranging from 9 to 784 dimensions.
Models with fewer than 25 dimensions have lower accuracy; all
models with more than 25 dimensions have $97\%$ or higher accuracy.  

\begin{figure}
  \centering
  \includegraphics[scale=.85]{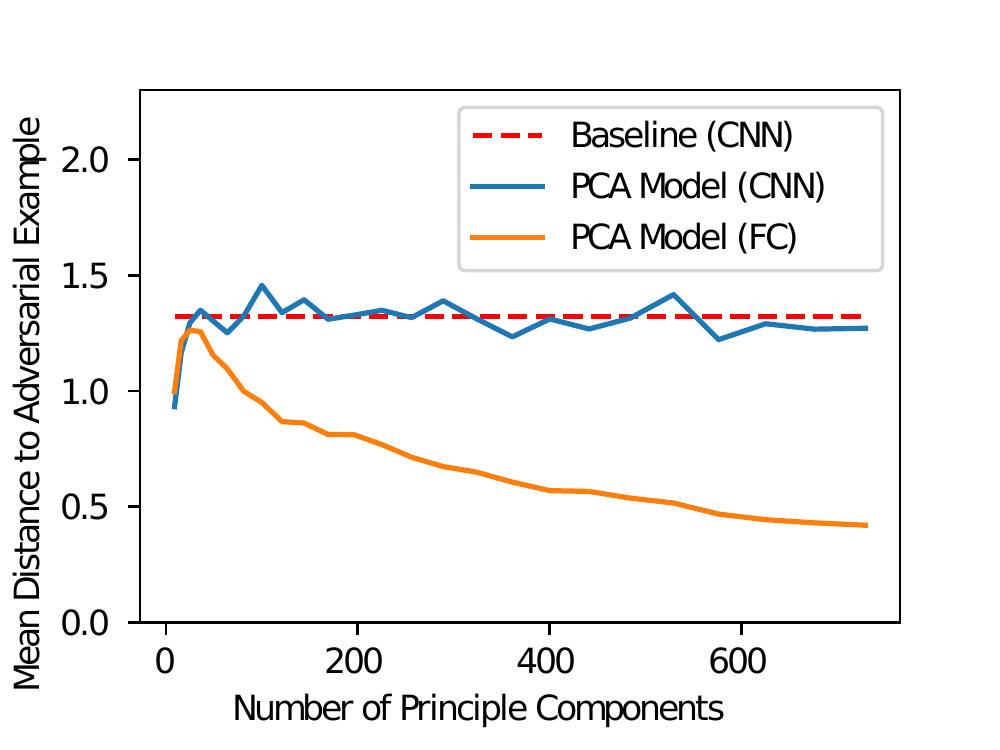}
  \caption{Performing dimensionality reduction increases the robustness of a
    100-100-10 fully-connected neural network, but is still less secure than
    just using an unsecured CNN (the baseline). Dimensionality reduction does not help on a
    network that is already convolutional.}
  \label{fig:dimreduction}
\end{figure}

\partb
We evaluate Bhagoji's defense by constructing targeted attacks
against all 26 models we trained.
We show the mean distortion for each model in Figure~\ref{fig:dimreduction}.
The most difficult model to attack uses
only the first 25 principal components; it is nearly $3\times$ more robust than
the model that keeps all 784 principal components.

However, crucially, we find that even the model that keeps the first 25 principal
components is \emph{less} robust than almost any standard, unsecured convolutional
neural network;
an unprotected network achieves both higher accuracy ($99.5\%$ accuracy)
and better robustness to adversarial examples (measured by the mean distortion).
In summary, Bhagoji's defense is not secure against white-box attacks.

\paragraph{Looking Deeper.}
Next, we show that this result is not an artifact of the network
architecture --- it is not caused just because
fully-connected network are less robust than convolutional networks.
We study a second algorithm that Bhagoji \emph{et al.} present but
did not end up using, which combines PCA with a convolutional neural network
architecture.
This allows us to perform an experiment where the network architecture
is held fixed, and the only change is whether dimensionality reduction
is used or not.
In particular, instead of using
the first $K$ principal components as features for a fully-connected network,
they use PCA to map each image into the reduced-dimensionality PCA
space, and then immediately map it back to the image space.
This effectively projects the image into a reduced-dimension manifold.
They train a convolutional classifier on the projected images.
This classifier achieves a higher accuracy
($99\%$ when using at least 25 dimensions).

In our experiments we found that this approach is no more robust
than an unsecured convolutional network (applied to the original image),
despite only using a limited number of the principal components.
We conclude
that defenses based on limiting the adversary to only the first principal
components are not effective.
It follows that Hendrycks \emph{et. al}'s defense is broken as well, as the
adversarial images generated in this way change only the first $K$ components
and leave all later components unchanged.

\subsection{Hidden Layer PCA}
\label{sec:pcacnn}

Li \emph{et al.} \cite{li2016adversarial} apply PCA to the values after inner convolutional layers
of the neural network, and use a cascade classifier to detect adversarial
examples. Specifically, they propose building a \emph{cascade
  classifier} that accepts the input as natural only if all classifiers $C_i$ accept
the input, but rejects it if any do. Each classifier $C_i$ a linear SVM
that acts on the PCA of the $i$th convolutional layer of the network.

They evaluate their scheme on ImageNet.  In the remainder of this section we
demonstrate their defense is not effective on MNIST and CIFAR;
others have shown that attacking
an ImageNet classifier would be even easier \cite{carlini2016towards,moosavi2016deepfool}.

\parta
Li \emph{et al.} generated adversarial examples with Szegedy \emph{et. al}'s
L-BFGS algorithm~\cite{szegedy2013intriguing}
and found that the first linear SVM achieved $80\%$ true positive
rate at $0\%$ false positive rate -- an ideal use-case for a cascade classifier.
We evaluated the effectiveness of their method at detecting adversarial
examples generated with C\&W's attack (when trained on adversarial examples
generated with the C\&W attack).
For MNIST, their scheme only achieves a $62\%$ true
positive rate at a $37\%$ false positive rate (and $56\%$ at $40\%$ on CIFAR)
on the first layer of the cascade classifier.
This results causes the cascade classifiers to have a $92\%$ and $98\%$
false positive rates on MNIST/CIFAR.
We conclude that Li's defense is not effective against the
C\&W attack.

\section{Distributional Detection}

Next, we study two defenses that detect adversarial examples by
comparing the distribution of natural images to the distribution of adversarial
examples.  They use classical statistical methods to distinguish natural images
from adversarial images.

\subsection{Maximum Mean Discrepancy}
\label{sec:mmd}

Grosse \emph{et al.} \cite{grosse2017statistical}
consider a very powerful threat model: assume we are given two sets of
images $S_1$ and $S_2$, such that we know $S_1$ contains only natural images, and
we know that $S_2$ contains either all adversarial examples, or all natural images.
They ask the question: can we determine which of these two situations is the case?

To achieve this, they use the Maximum Mean Discrepancy (MMD) test
\cite{borgwardt2006integrating,gretton2012kernel},
a statistical hypothesis test that answers the question ``are
these two sets drawn from the same underlying distribution?''


The MMD is a theoretically useful technique that can be formally shown to
always detect a difference if one occurs. However, it is computationally infeasible
to compute, so a simple polynomial-time approximation is almost always used.
In our experiments, we use the same approximation used by
Grosse \emph{et al.}~\cite{gretton2012kernel}.

To test whether $X_1$ and $X_2$ are drawn from 
the same distribution,
Grosse \emph{et al.} use Fisher's permutation test \cite{oden1975arguments}
with the MMD test statistic.
To do this, initially let $a=MMD(X_1,X_2)$. Then, shuffle
the elements of $X_1$ and $X_2$ into two new sets $Y_1$ and $Y_2$,
and let $b=MMD(Y_1,Y_2)$. If $a<b$ then reject the null hypothesis, and
conclude they are drawn from different distributions.
Repeat this process many times, and repor the p-value as the fraction of
times the null hypothesis is rejected.



As a first step, we verified that MMD correctly detects natural instances as being natural.
To do this, we assign $X_1$ and $X_2$ to be random samples from the test set.
The procedure above correctly fails to reject the
null hypothesis ($p > 0.05$) in distinguishing the two distributions.

\parta
We repeated this experiment, this time producing targeted adversarial
examples with C\&W's attack algorithm. Even when using a set of $100$ images, MMD
fails to reject the null hypothesis ($p>0.05$).
MMD also fails to detect attacks on CIFAR,
even with a set size as large as $10,000$ (which uses every sample
in the test set).
Since MMD is one of the most powerful multidimensional statistical tests,
and even it is not effective, we argue that without significant modification,
statistical tests will not be able to detect adversarial examples.



\subsection{Kernel Density Estimation}
\label{sec:de}

Feinman \emph{et al.} \cite{feinman2017detecting} present a defense they call \emph{kernel density estimation}.
They use a Gaussian Mixture Model to model outputs from the final hidden layer of a neural
network, and argue that adversarial examples belong to a different distribution
than that of natural images.

Specifically, given an instance $x$ classified as label $t$,
kernel density estimation estimates the likelihood of $x$ as
\[ KDE(x) = {1 \over |X_t|} \sum\limits_{s \in X_t} \exp({|F^{n-1}(x)-F^{n-1}(s)|^2 \over \sigma^2}) \]
where $X_t$ is the set of training instances with label $t$
and $F^{n-1}(x)$ is the output of the final hidden layer on input $x$.
The detector is therefore
constructed by selecting a threshold $\tau$ and reporting $x$ as
adversarial if $KDE(x) < \tau$, otherwise reporting $x$ as natural.

The motivation behind this approach is that the later hidden layers of a neural
network have been shown to capture high-level semantic information about the input.
Therefore, using a simple classifier on this final layer will be more accurate than
if it were applied to the original input images, as the prior defense did.

\parta
Feinman's defense is able to detect C\&W adversarial examples on MNIST, but not
on CIFAR.
Looking deeper, on CIFAR, for
each image in the test set $x$ and closest adversarial
example $x'$, we compare $KDE(x')$ to $KDE(x)$.
Surprisingly, we find that $80\%$ of the time, the adversarial example
has a \emph{higher} likelihood score than the original image.
Therefore, Feinman's defense cannot work on CIFAR. 
In the remainder of this section, we show how to break this defense on MNIST
with increased distortion.

\partb
To mount a white-box attack, we construct a new minimization formulation that
differs from the original only in that we introduce a new loss term
$\ell_2(x')$ that penalizes being detected by the detector:
\[ \text{minimize } \; \|x-x'\|_2^2 + c \cdot (\ell(x') + \ell_2(x')) \]
where we define
\[ \ell_2(x') = \max(-\log(KDE(x'))-\epsilon,0) \]
where $\epsilon$ controls the likelihood measure of the adversarial examples.
In our attack, we set $\epsilon$ to the median of $-\log(KDE(\cdot))$ on the training
set, so that $\ell_2(x') \le 0$ if and only if $KDE(x')$ is greater than half of the
training instances KDE.

In practice, we mount this attack in two phases. First, we solve the original
C\&W minimization formulation to obtain an adversarial example $\hat x$.
Typically $\hat x$ will be detected by the detector, so in the second phase
we modify it to no longer be detected: we use this $\hat x$ as the initial value
of $x'$ in the above optimization problem and use gradient descent to
improve it. Performing this two-step optimization is
useful to allow for different constants $c$ chosen for initially generating an
adversarial example and for making it not detected.

This approach is effective at generating adversarial examples for
MNIST that are not detected by Feinman's defense.
The mean distortion is $3.81$, up from $2.05$ on an unsecured network.
In order to implement this attack more efficiently, instead of computing the KDE
across all training instances during the inner optimization loop, we randomly
sample $100$ instances ($<5\%$ of the total) at first and use only those. A distortion
of $3.81$ begins to allow humans to notice the adversarial nature of these
images (Figure~\ref{fig:show_mnistcifar}).

We believe the reason this defense is so much more effective on MNIST than
on CIFAR because 
this defense is effectively a K-nearest-neighbor
classifier with a neural network feature extractor.
In fact, prior work has shown that using an RBF kernel SVM (essentially identical
to what is done here) is effective at defending MNIST to adversarial examples. 

\partc
Again we only demonstrate the vulnerability on MNIST: the defense does not
work on CIFAR.
Note that our white-box attack algorithm didn't use the entire training set;
it uses only a random sample of 100 instances from the training set
(less than $5\%$ of the training instances).
This hints that black-box attacks might succeed.
This is indeed the case: we found that when
the defender uses the first $95\%$ of the training
data, and the last $5\%$ of the training data is left for the adversary (who uses
it to form their own KDE estimator), the attack still succeeds with equal distortion.

\section{Normalization Detection}

\subsection{Dropout Randomization}
\label{sec:dropout}

Feinman \emph{et al.} propose a second detection method called
\emph{Bayesian neural network uncertainty} that measures the uncertainty of
the neural network on the given input. Instead of relying on the reported
confidence of the network (which can easily be controlled by an adversary),
they add randomization to the network.
The hope is that a natural image will have the
same (correct) label regardless of the random values chosen, while adversarial
examples won't always be predicted with the same label.
\emph{Dropout} \cite{srivastava2014dropout} is used as the method of adding randomness.


To determine the uncertainty of the network on a given instance, we apply
the randomized network $F_r$ (with dropout enabled) to the input $L$ times.
Then, the uncertainty is defined as
  \[ U(x) = \left({1 \over L} \sum\limits_{i=1}^L \|F_r(x)\| \right) -
  \left\| {1 \over L} \sum\limits_{i=1}^L F_r(x) \right\| \]
where we take $\|y\|$ to mean $y^T y$ (the squared $L_2$ norm). Notice that if
the predictions $F_r(x)$ are the same every time, the uncertainty is $0$.
In other words, the uncertainty is the sum of the variances of each
component of the output,
computed over the $L$ outputs from the randomized network.

To classify an input $x$ as adversarial or natural, we pick a threshold $\tau$ and
a number of times to evaluate the network $L$ and
compute $U(x)$. We have found the algorithm is not sensitive to the choice of $L$
as long as it is greater than 20 (and we pick $L=30$ for the remainder of this section).
We return ``adversarial'' if $U(x) > \tau$ and ``natural'' otherwise.

When we run this detection scheme on the adversarial examples generated by the
FGS and JSMA attack algorithms on MNIST, it detects $96\%$ of adversarial
examples with a false positive rate lower than $1\%$ when $\tau=0.02$.

\parta
We generated adversarial examples with C\&W's attack and find
$75\%$ of these adversarial examples are detected by this defense on MNIST.
Surprisingly, this defense is even more effective at detecting attacks on CIFAR:
it detects $95\%$ of C\&W adversarial examples.

\partb
Our white-box attacks work by defining an objective function that combines
the model and detector.
In an initial attempt,  we follow the approach taken to break the prior detection schemes, let
\[ G(x)_i = \begin{cases}
  Z(x)_i & \text{if } i \le N \\
  (1+U(x)-\tau) \cdot \max\limits_i Z(x)_i & \text{if } i = N+1
\end{cases} \]
(where $N+1$ is the new adversarial class).
This objective function is defined so that
if $U(x) > \tau$ (i.e., $x$ is classified as adversarial by Feinman's defense) then we will have
\[\text{arg max}_i(G(x)_i)=N+1,\]
and if $U(x) < \tau$ (i.e., $x$ is classified as natural) then
\[ \text{arg max}_i(G(x)_i) = \text{arg max}_i(F(x)_i).\]

We then attempt to generated adversarial examples by applying C\&W's attack
to $G(\cdot)$.
However, this attack fails: the defense is still able to detect $60\%$ of adversarial
examples.

\paragraph{Looking Deeper.}
We investigate why this attack failed.
Because values of the function $U(\cdot)$ are computed by
calling $F(\cdot)$ (and not $Z(\cdot)$), there is no good constant
$c$ that C\&W's algorithm can select. This causes gradient descent to perform
poorly and rarely returns successful adversarial examples.

Therefore, for this defense, we abandon the approach of constructing a single
function to optimize over and instead change the loss function $\ell$ used by
the attack. Recall that the typical loss function used (for a given network $Z(\cdot)$)
is
\[\ell_Z(x') =  \max(\max \{ Z(x')_i : i \ne t\} - Z(x')_t, -\kappa). \]
If every prediction $F_r(x)$ of the randomized network
produced a very similar result, the uncertainty value $U(x)$ would be very small.

We sample $K$ different deterministic networks $\{Z_j(\cdot) : j \in [1,K]\}$ each with
different randomness used during dropout. If we were able to have $\text{arg max}_i Z_j(x)_i=t$
for every network $j$, for $K$ big enough, it would be highly likely that $F_r(x)$
would always produce label $t$ for any randomness. Thus, we construct a new loss function
$\ell'(x') = \sum_{j=1}^K \ell_{Z_j}(x')$
as the average of the loss functions on each fixed model $Z_j$.
Then we use C\&W's attack with this revised loss function.

This approach successfully generates adversarial examples that fool
the dropout defense with $98\%$ success.
On MNIST, the mean $l_2$ distortion is $3.68$, up from the baseline of $2.05$.
This is the largest distortion required by any of the defenses we have
evaluated; the distortion here is nearing the levels of human perceptibility
(Figure~\ref{fig:show_mnistcifar}).
On CIFAR the distortion required again increases to $1.1$, a factor of $5\times$
larger, but is still entirely imperceptible (Figure~\ref{fig:show_mnistcifar}).

\partc
It turns out that adversarial examples generated with the
white-box approach transfer
with high probability across models. This is due to the fact that our white-box attack
assumes we do not know the exact randomization settings, and therefore construct
adversarial examples that are effective regardless of randomization. This is 
similar to the black-box threat model, where the adversary does not have access
to the model parameters.

However, to improve the rate of transferability, 
we again construct two models $F(\cdot)$ and $G(\cdot)$ on
different subsets of the training
data. We provide the adversary access to the parameters of $F$ and use
the white-box attack above to generate adversarial examples for $F$; we then
test whether they fool $G$.

On MNIST, we find that adversarial examples generated with our algorithm
transfer to $G$ with
$90\%$ success.
We can increase the transfer rate to $98\%$ at the cost of increasing
the mean distortion only $15\%$, to $4.23$.
While this $15\%$ increase may seem like a failure of this defense under the
black-box threat model, we see this instead as a success of
defense under the white-box threat model.
It has made constructing adaptive white-box attacks nearly as hard as constructing
black-box attacks.

In fact, we find that this is the reason why the CIFAR network has such a
larger increase in distortion: to obtain transferable cifar adversarial
examples requires a distortion of approximately $4\times$ larger
than non-transferable attacks. We consider this the only defense we
evaluate that is
not completely broken (even though we still can generate adversarial examples
that are imperceptible on CIFAR).


\subsection{Mean Blur}
\label{sec:blur}

The second detection method proposed by Li \emph{et al.} 
applies a
$3 \times 3$ average filter to blur the image before applying the classifier.
The authors admit this defense is ``overly simplistic''
but still argue it is effective at alleviating adversarial examples.
We confirm this simple defense can remove adversarial examples generated with
fast gradient sign, as they found in their paper.

\parta When we use C\&W's attack, we find that this defense effectively
removes low-confidence adversarial examples: $80\%$ of adversarial examples
(at a mean $L_2$ distortion of $2.05$) are no longer classified incorrectly.

This attack can even partially alleviate high-confidence adversarial examples.
To ensure they remain adversarial after blurring, we must increase the distortion by a
factor of $3\times$.

\partb
Observe that taking the mean over every
$3 \times 3$ region on the image is the same as adding another convolutional layer
to the beginning of the neural network with one output channel that performs this
calculation.
Given the network $F$, we define $F'(x) = F(\text{blur}(x))$
and apply C\&W's attack against $F'$. When we do so, we find that the
mean distance to adversarial examples does not increase.
Therefore, blurring is not an effective defense.

\section{Lessons}

\subsection{Properties of adversarial examples}

After examining these ten defenses, we now draw conclusions about the
nature of the space of adversarial examples and the ability to
detect them with different approaches.

\paragraph{\textbf{Randomization can increase required distortion.}}
By far the most effective defense technique, dropout randomization, made
generating adversarial examples nearly five times more difficult on CIFAR. In
particular, it makes generating adversarial examples on the network as difficult
as generating transferable adversarial examples, a task known to be harder
\cite{papernot2016transferability}.
Additionally, 
if it were possible to find a way to eliminate transferability,
a randomization-based defense may be able to detect adversarial examples.
At this time, we believe this is the most promising direction of future work.

\paragraph{\textbf{MNIST properties may not hold on CIFAR}} 
Most defenses that increased the distortion on MNIST had a significantly
lower distortion increase on CIFAR. In particular, kernel density estimation,
the most effective defense on MNIST, was completely ineffective on CIFAR.

\paragraph{\textbf{Detection neural networks can be bypassed.}}
Across all of the defenses we evaluate, the least effective schemes used another
neural network (or more neural network layers) to attempt to identify adversarial
examples. Given that adversarial examples can fool a single classifier, it makes
sense that adversarial examples can fool a classifier and detector.

\paragraph{\textbf{Operating on raw pixel values is ineffective.}}
Defenses that operated directly on the pixel values were too
simple to succeed. On MNIST, these defenses provided reasonable robustness
against weak attacks; however when evaluating on stronger attacks,
these defenses all failed. This should not be surprising: the reason neural networks are
used is that they are able to extract deep and meaningful features from the
input data. A simple linear detector is not effective at classification when
operating on raw pixel values, so it should not be surprising it does not work
at detecting adversarial examples. (This can be seen especially well on CIFAR,
where even weak attacks often succeed against defenses that operate on the
input pixel space.)


\subsection{Recommendations for Defenses}

We have several recommendations for how researchers proposing new defenses
can better evaluate their proposals. Many of these recommendations may appear
to be obvious, however most of the papers we evaluate do not follow any.

\paragraph{\textbf{Evaluate using a strong attack.}}
Evaluate proposed defenses using the strongest attacks known.
\emph{Do not use fast gradient sign or JSMA exclusively}:
most defenses that detect these attacks fail against stronger attacks.
In particular, Fast gradient sign was not even designed to produce
high-quality attacks: it was created to demonstrate
neural networks are highly linear.
Using these algorithms as a first test is reasonable first step,
but is not sufficient.
We recommend new schemes evaluate against strong iterative attacks.

\paragraph{\textbf{Demonstrate white-box attacks fail.}} It is not sufficient to show that
a defense can detect adversarial examples: one must also show that a
adversary
aware of the defense can not generate attacks that evade detection.
We show how to perform that kind of evaluation:
construct a differentiable function that is minimized when the image
fools the classifier and is treated as natural by the detector,
and apply a strong iterative attack (e.g., C\&W's
attack) to this function.




\paragraph{\textbf{Report false positive and true positive rates.}}
When constructing a detection-based defense, it is not enough to report
the accuracy of the detector.
A $60\%$ accuracy can either be very useful
(e.g., if it achieves a high true-positive rate at a $0\%$ false-positive rate) or
entirely useless (e.g., if it detects most adversarial images as adversarial at the
cost of many natural images as adversarial).
Instead, report both the false positive and true positive rates.
To allow for comparisons with other work, we suggest reporting at least the true positive rate
at $1\%$ false positive rate;
showing a ROC curve would be even better.

\paragraph{\textbf{Evaluate on more than MNIST}} We have found that defenses that only
evaluated on the MNIST dataset typically either (a) were unable to produce
an accurate classifier on CIFAR, (b) were entirely useless on CIFAR and were not able
to detect even the fast gradient sign attack, or (c) were even weaker against attack on CIFAR than the other defenses we evaluated.
Future schemes need to be evaluated on multiple data sets --- evaluating
their security solely on MNIST is not sufficient.
While we have found CIFAR to be a reasonable task for evaluating
security, in the future as defenses improve
it may become necessary to evaluate on harder
datasets (such as ImageNet \cite{deng2009imagenet}).

\paragraph{\textbf{Release source code.}}
In order to allow others to build on their work,
authors should release the source code of their defenses.
Not releasing source code only sets back the
research community and hinders future security analysis.
Seven of the \numdefenses we evaluate did not
release their code (even after contacting the authors),
requiring us to reimplement the defenses before evaluation.

\section{Conclusion}

Unlike standard machine-learning tasks, where achieving a higher accuracy
on a single benchmark is in itself a useful and interesting result,
this is not sufficient for secure machine learning.
We must consider how an attacker might react to any proposed defense,
and evaluate whether the defense remains secure against an attacker
who knows how the defense works.

In this paper we evaluate ten proposed defenses and
demonstrate that none of them
are able to withstand a white-box attack. We do this by constructing
defense-specific loss functions that we minimize with a strong iterative
attack algorithm.
With these attacks, on CIFAR an adversary can create imperceptible
adversarial examples for each defense.

By studying these \numdefenses defenses, we have
drawn two lessons: existing defenses lack thorough security evaluations, and
adversarial examples are much more difficult to detect than
previously recognized.
We hope that our work will help raise the bar for evaluation of proposed
defenses and perhaps help others to construct more effective defenses.
Further, our evaluations of these defenses expand on what is believed to be
possible with constructing adversarial examples: we have shown that, so far,
there are no known intrinsic properties that differentiate adversarial examples
from regular images.
We believe that constructing defenses to adversarial examples is an important
challenge that must be overcome before these networks are used in potentially
security-critical domains, and hope our work can bring us closer towards this goal.

\section{Acknowledgements}
We would like to thank Kathrin Grosse, Reuben Feinman, Fuxin Li, and Metzen Jan Hendrik
for discussing their defenses with us, along with the anonymous reviewers for their feedback.
This work was supported by the AFOSR under MURI award FA9550-12-1-0040, Intel through the ISTC for Secure Computing, the Hewlett Foundation through the Center for Long-Term Cybersecurity, and Qualcomm.

{\footnotesize
\bibliographystyle{ACM-Reference-Format}
\bibliography{paper}
}

\end{document}